# IdiomX: A Multilingual Benchmark for Idiom Understanding, Retrieval, and Semantic Interpretation

**Ayman Ali Sharara**
MSc Data Science & Machine Learning (SPOC S21)
Data ScienceTech Institute (DSTI)
Email: ayman.sharara@edu.dsti.institute

**Co-author: Hanna Abi Akl**
Email: hanna.abi-akl@dsti.institute

**Abstract**

Idiomatic expressions remain a persistent challenge for natural language processing because their meanings are often non-compositional, context-dependent, and difficult to align across languages. Existing idiom resources are often limited in scale, contextual diversity, or multilingual coverage, restricting their utility for modern language models.

We introduce IdiomX, a large-scale multilingual benchmark for idiom understanding, retrieval, and interpretation, constructed through a reproducible multi-stage pipeline combining lexical resource extraction, large-scale normalization, controlled large language model enrichment, and structured validation. The resulting dataset contains over 190K contextualized examples spanning more than 12K idioms, with aligned English, Arabic, and French semantic representations, idiomatic and literal usage labels, and rich linguistic metadata.

Building on this resource, we define a unified four-task benchmark covering (1) idiom detection, (2) context-to-idiom retrieval, (3) Arabic-to-English idiom retrieval, and (4) idiom interpretation, extending evaluation from figurative recognition to semantic grounding and explainable meaning retrieval.

Experiments show that contextual transformer models substantially improve idiom detection, while hybrid retrieval and reranking architectures significantly strengthen both monolingual and cross-lingual idiom retrieval. Results further demonstrate that idiom interpretation can be effectively modeled as a semantic retrieval task, introducing interpretability as a complementary benchmark dimension.

Overall, IdiomX provides a scalable benchmark for studying idiomatic language as a progression from detection to retrieval and semantic interpretation, and offers a modular framework extensible to additional languages and figurative reasoning tasks.

## I. Introduction

Idiomatic expressions remain a major challenge for natural language understanding due to their non-compositional semantics, where meaning cannot be inferred directly from constituent words. Expressions such as *kick the bucket* or *spill the beans* require contextual and cultural reasoning beyond literal analysis, making idioms difficult for systems performing semantic interpretation, machine translation, information retrieval, and multilingual reasoning.

Despite major advances in contextual embeddings and large language models (LLMs) [6], idiomatic language continues to be a persistent source of errors. Models often struggle to distinguish literal from figurative usage, retrieve idiomatic meaning from context, or align idiomatic semantics across languages. A major bottleneck is data. Existing resources such as VNC-Tokens [1], SemEval-2013 Task 5b [2], PIE [3], and MAGPIE [4] have provided important benchmarks, but remain limited in scale, contextual diversity, multilingual coverage, or support for broader idiom reasoning tasks.

To address these limitations, we introduce IdiomX, a large-scale multilingual benchmark for idiom understanding, retrieval, and interpretation. IdiomX contains over 190K contextualized examples spanning more than 12K idioms, with idiomatic and literal usage labels, aligned English–Arabic–French semantic representations, and rich linguistic metadata. The dataset is constructed through a reproducible multi-stage pipeline combining lexical resource extraction, large-scale normalization, controlled LLM enrichment, and structured validation, supported by semantic consistency checks, deduplication, and leakage-safe evaluation splits.

Beyond dataset construction, IdiomX defines a unified benchmark spanning four complementary tasks: **(1)** idiom detection, **(2)** context-to-idiom retrieval, **(3)** Arabic-to-English idiom retrieval, and **(4)** idiom interpretation, which focuses on retrieving and explaining idiomatic meaning across languages. Together, these tasks form a progression from recognition to retrieval and semantic interpretation, enabling a broader evaluation of figurative language understanding than prior benchmarks.

Experimental results show strong performance using transformer classifiers, hybrid dense-lexical retrieval, multilingual embedding models, and interpretation-oriented retrieval methods, while also highlighting persistent challenges in contextual ambiguity and cross-lingual idiomatic semantics.

The main contributions of this work are:

1. **IdiomX**, a large-scale multilingual benchmark for idiom understanding with aligned semantic representations across English, Arabic, and French.
2. A reproducible multi-stage data construction and enrichment pipeline for generating high-quality idiomatic examples with structured validation.
3. A unified benchmark spanning four tasks covering idiom detection, retrieval, cross-lingual retrieval, and idiom interpretation.
4. Extensive experiments and analyses establishing strong baselines and demonstrating IdiomX as a scalable benchmark for figurative language research.

Overall, IdiomX bridges dataset construction and task-oriented evaluation, providing a foundation for research on figurative reasoning, multilingual semantic retrieval, and idiom interpretation.

## II. Related Work

### A. *Existing Idiom Datasets*

Research on idiomatic language understanding has produced several benchmark datasets for modeling figurative language. Early resources such as VNC-Tokens [1] and SemEval-2013 Task 5b [2] established foundational benchmarks for distinguishing idiomatic and literal usage, while larger datasets such as PIE [3] and MAGPIE [4] expanded contextual coverage and scale. Despite these advances, most existing resources remain predominantly monolingual, focus on a single task (typically idiom detection), and provide limited support for multilingual semantic alignment or unified multi-task evaluation.

More recent work has explored multilingual and cross-lingual idiom resources [5], [6], yet these often rely on direct translation pairs and lack large-scale contextual examples, balanced idiomatic versus literal supervision, and broader task coverage. These limitations motivate the need for scalable benchmarks that support richer idiom understanding across languages and tasks.

### B. *Modeling Approaches for Idiom Understanding*

Computational approaches for idiom understanding have evolved from feature-based statistical methods toward neural contextual models. Traditional approaches based on lexical features and shallow classifiers were limited in capturing non-compositional meaning. Recent transformer-based models such as BERT [13] and multilingual representation models [6], together with dense retrieval methods based on Sentence-BERT [17], have significantly improved idiom detection and semantic retrieval.

However, model performance remains strongly dependent on dataset quality, contextual diversity, and benchmark design. This makes large-scale, well-structured resources essential for robust evaluation and motivates the development of comprehensive benchmarks such as IdiomX.

### C. *IdiomX in Context*

As shown in Table 1, IdiomX substantially extends prior idiom resources in scale, multilingual coverage, and support for a unified four-task benchmark. Unlike existing datasets that primarily target single-task idiom detection, IdiomX supports unified evaluation across idiom detection, context-to-idiom retrieval, cross-lingual retrieval, and idiom interpretation.

| Dataset | Lang | Idioms | Inst. | Bi | Ref |
|---|---|---|---|---|---|
| VNC-Tokens | EN | 53 | 2,984 | N | [1] |
| SemEval-2013 Task 5b | EN | 85 | ~4,350 | N | [2] |
| IDIX | EN | 52 | 4,022 | N | [20] |
| MAGPIE | EN | ~1,700 | 56,000 | N | [4] |
| PIE-English | EN | ~1,200 | ~20,100 | N | [21] |
| IdiomX | EN AR FR | 12k | 190k | Y | [23] |

Table 1: Comparison of IdiomX with existing idiom datasets.

This broader coverage positions IdiomX as a unified benchmark for evaluating idiomatic language understanding across multiple tasks, languages, and semantic interpretation settings.

## III. DataSet Construction

The IdiomX dataset is constructed through a structured multi-stage pipeline that integrates multiple data sources, controlled LLM-based enrichment, and systematic validation. Unlike conventional single-source approaches, IdiomX is built through three complementary pipelines targeting: (1) curated lexical idioms, (2) modern and informal idiomatic usage, and (3) synthetically generated idioms. These pipelines are merged, deduplicated, and validated to form a unified high-quality dataset.

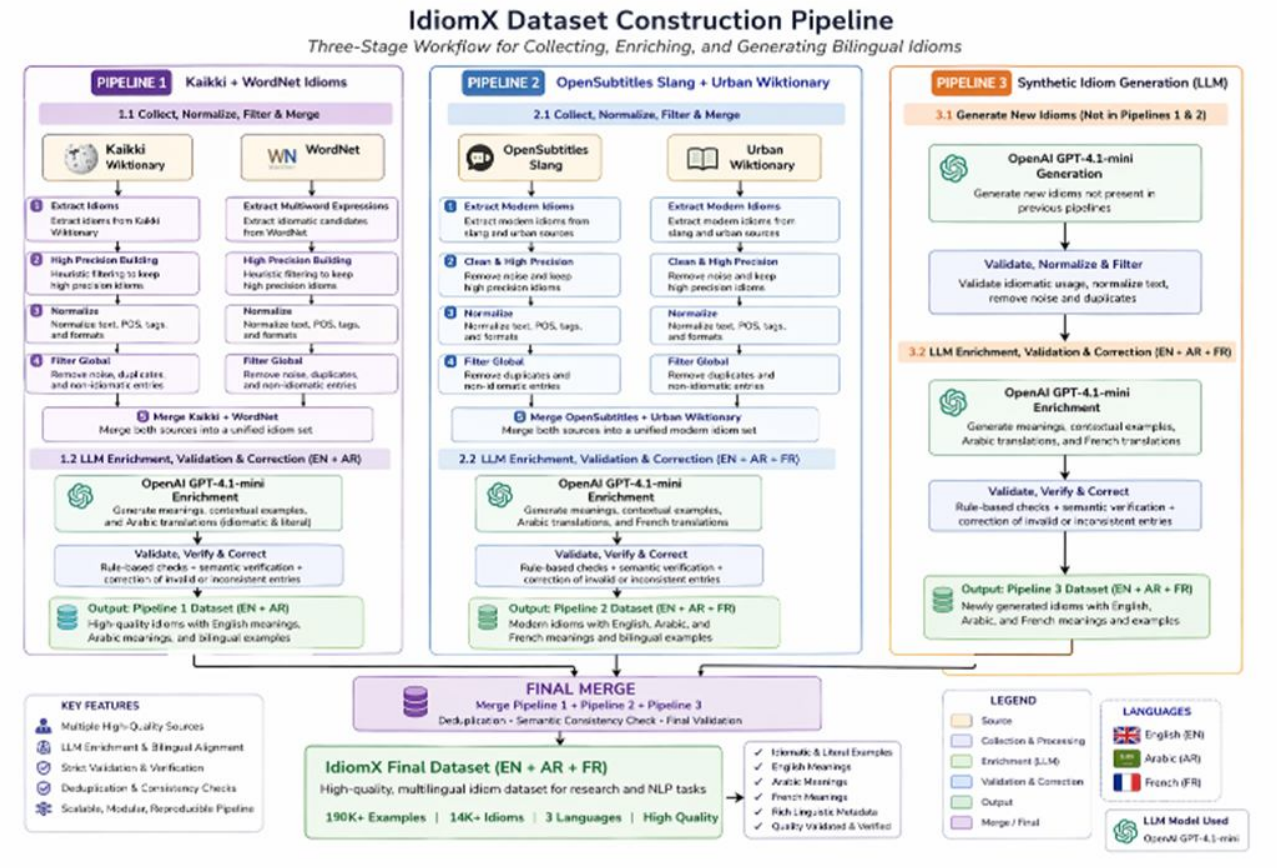

Figure 1 illustrates the overall dataset construction pipeline.

### A. *Multi-Stage Data Construction Pipeline*

IdiomX combines three complementary sub-pipelines:

**1) Core Idioms Pipeline (Kaikki + WordNet)**

This pipeline extracts idioms from structured lexical resources and applies filtering, normalization, and merging to construct a high-precision idiom inventory.

**2) Modern Idioms and Slang Pipeline (OpenSubtitles + Urban + Wiktionary)**

To capture contemporary usage, modern idioms and slang are collected from informal and conversational sources, then normalized and merged through the same quality-controlled pipeline.

**3) Synthetic Idiom Generation Pipeline (LLM-based)**

To extend coverage beyond existing resources, additional idioms are generated using GPT-4-class models [24], followed by filtering, normalization, and deduplication.

After independent processing, outputs from all pipelines are merged into a unified canonical idiom representation.

### *B. LLM Enrichment, Validation, and Quality Control*

Following collection, idioms undergo structured enrichment using GPT-4-class models [24] through a scalable batch-processing pipeline. For each idiom, the system generates contextual examples, idiomatic and literal usage labels, semantic explanations, multilingual meanings (English, Arabic, French), and adversarial examples for harder evaluation cases.

Quality assurance combines structured prompt constraints, automated semantic verification, rule-based correction, and statistical validation [26]. Generated instances are automatically verified, corrected, or rejected based on linguistic quality, semantic alignment, and structural completeness. This multi-layer quality-control framework ensures both diversity and consistency while reducing noise at scale.

### *C. Example Dataset Instances*

**Example idiomatic usage**
"He finally kicked the bucket after a long illness."
Meaning: To die.

**Example literal usage**
"She kicked the bucket across the yard."

**Rejected low-quality example**
"He kicked bucket yesterday very sadness."
This example is rejected due to grammatical and semantic errors.

### *D. Final Dataset Characteristics*

The final IdiomX release contains over 190k contextualized examples covering more than 12k idioms, with aligned multilingual semantic representations and balanced idiomatic versus literal supervision. This design supports unified evaluation across detection, retrieval, cross-lingual mapping, and idiom interpretation tasks.

Leakage-safe train/test splitting is applied to support reliable benchmark evaluation and reduce memorization bias.

## IV. Dataset Analysis

We analyze IdiomX to evaluate its linguistic diversity, semantic quality, and suitability for idiom understanding benchmarks. The analysis focuses on the full dataset.

### *A.* Linguistic and Semantic Properties

The dataset contains over 190K contextualized examples covering more than 12K idioms, with substantial lexical and contextual diversity. The compositionality distribution shows that most idioms are semi-opaque, followed by opaque and transparent expressions. This confirms that IdiomX contains expressions across different levels of semantic complexity, requiring models to handle both partially interpretable and highly non-compositional meanings.

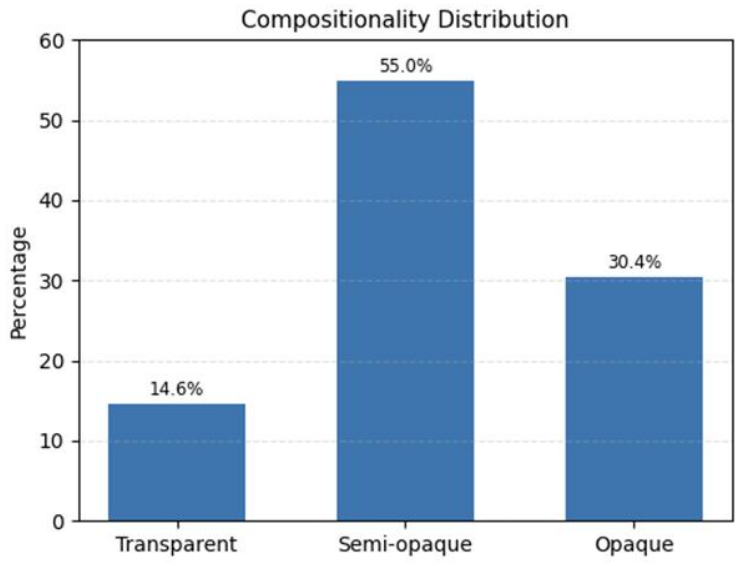

Figure 2: compositionality distribution.

### *B.* Data Quality and Label Balance

IdiomX exhibits minimal duplication, with a reuse factor close to 1.04, reducing memorization risk and supporting robust evaluation. The semantic quality distribution shows that most examples are high quality, confirming the effectiveness of the enrichment and validation pipeline.

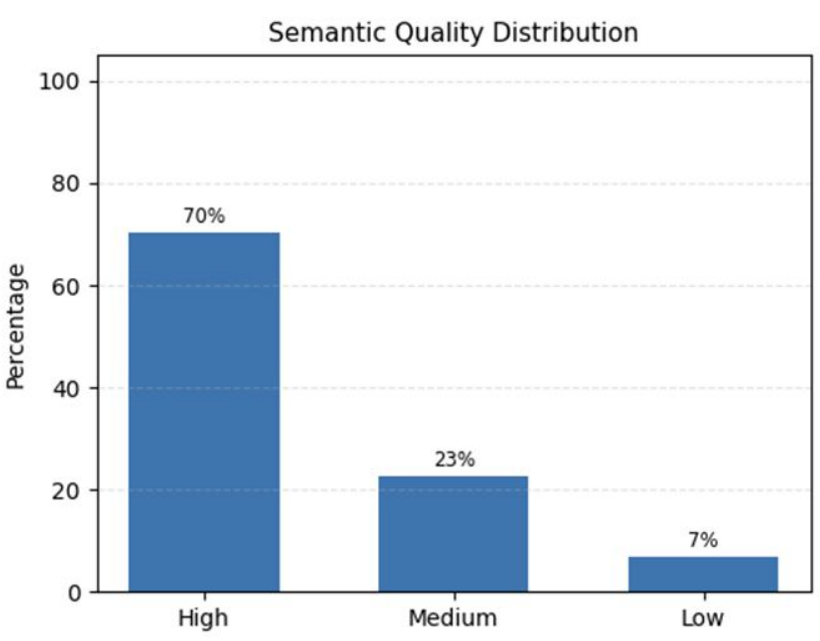

Figure 3: semantic quality distribution.

The dataset also maintains a near-balanced distribution between idiomatic and literal examples, with a smaller proportion of borderline cases. This balance is important for evaluating models that must distinguish figurative meaning from literal usage under realistic ambiguity.

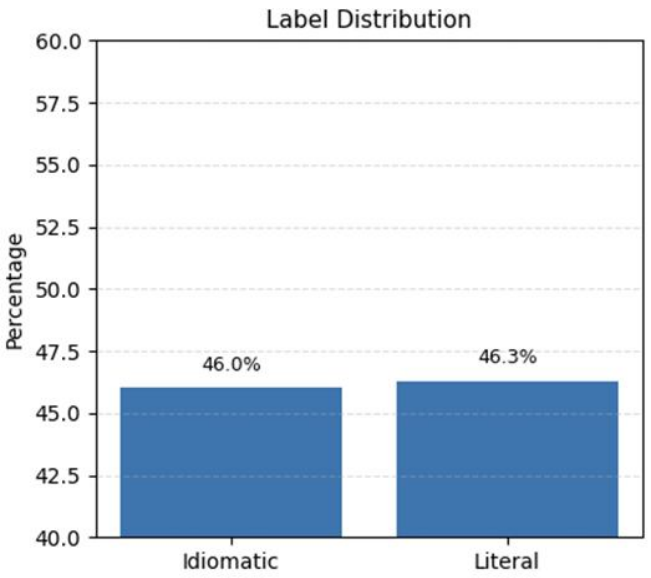

Figure 4: Label Distribution.

### *C. Dataset Release*

The final dataset is released as a single consolidated multilingual resource with aligned English, Arabic, and French semantic representations. It supports the four IdiomX benchmark tasks: idiom detection, context-to-idiom retrieval, Arabic-to-English idiom retrieval, and idiom interpretation.

## V. Benchmark Tasks

To comprehensively evaluate idiom understanding, we define four complementary benchmark tasks spanning figurative disambiguation, semantic retrieval, cross-lingual alignment, and idiom interpretation.

Together, these tasks form a unified evaluation framework that moves beyond isolated idiom detection benchmarks and enables systematic analysis of idiomatic language understanding across multiple levels: recognizing idiomatic usage, retrieving canonical idioms from context, aligning idioms across languages, and interpreting idiomatic meaning.

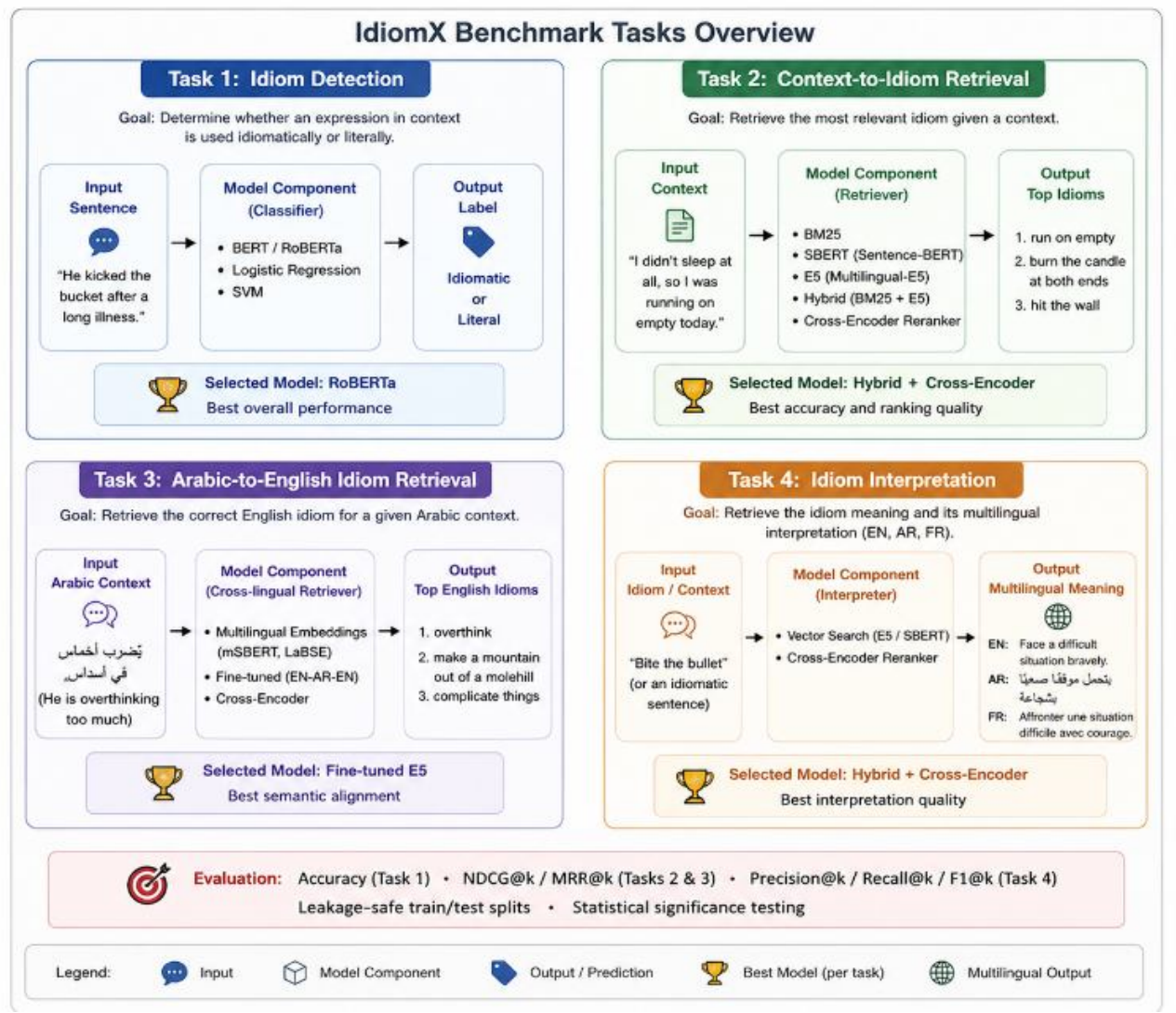

Figure 5: Unified four-task benchmark framework of IdiomX

### *A.* Task 1: Idiom Detection

Idiom detection focuses on identifying whether an expression in a sentence is used idiomatically or literally, a well-known challenge in figurative language understanding [1], [2].

We formulate this task as a binary classification problem.

Each instance consists of:

- **Input:** a sentence containing a target expression
- **Output:** a label ∈ {idiomatic, literal}

The task evaluates a model's ability to distinguish figurative meaning from literal interpretation, particularly in cases where the same expression can exhibit dual meanings depending on context (e.g., "kick the bucket").

This task primarily measures contextual disambiguation capability, which is essential for downstream applications such as machine translation, semantic parsing, and dialogue systems.

### *B.* Task 2: Context-to-Idiom Retrieval

Given a contextual sentence, the goal is to retrieve the idiom that best captures the underlying meaning expressed in the context.

We formulate this as a closed-set retrieval problem over a predefined idiom inventory.

Each instance consists of:

- **Input:** a contextual sentence
- **Output:** a ranked list of candidate idioms

The objective is to rank the correct idiom at the top of the candidate list.

This task evaluates deep semantic understanding and expressive alignment, requiring models to map contextual meaning to an appropriate figurative expression rather than relying on surface-level lexical cues.

### *C.* Task 3: Arabic-to-English Idiom Retrieval

This task extends retrieval to a cross-lingual setting, where the input is an Arabic sentence and the goal is to retrieve the corresponding English idiom.

We formulate this as a ranking problem, where candidate idioms are scored based on their semantic similarity to the input.

Each instance consists of:

- **Input:** An Arabic contextual sentence
- **Output:** a ranked list of English idioms

This task evaluates a model's ability to perform cross-lingual semantic alignment, particularly for figurative expressions that often lack direct lexical correspondence [6].

It is especially challenging due to differences in cultural context, expression structure, and semantic abstraction across languages.

### *D.* Task 4: Idiom Interpretation and Meaning Retrieval

We formulate this task as a semantic retrieval and interpretation task over a canonical idiom inventory.

Task 4 focuses on interpreting an idiom by retrieving its intended meaning and multilingual semantic explanation. Given an idiom or an idiomatic sentence, the system retrieves the most relevant canonical idiom and returns its meaning in English, Arabic, and French when available.

Each instance consists of:

- Input: an idiom or idiomatic sentence
- Output: a ranked list of candidate idioms with multilingual meanings

This task evaluates whether a model can move beyond detecting or retrieving an idiom and instead provide interpretable semantic grounding. It is particularly useful for applications such as language learning, machine translation support, multilingual search, and explainable AI systems.

### *E.* Unified Evaluation Perspective

Together, these tasks form a progressive evaluation framework:

- **Task 1**: Figurative usage recognition
- **Task 2:** Context → idiom retrieval
- **Task 3:** Cross-lingual idiom retrieval
- **Task 4:** Idiom → multilingual meaning interpretation

This progression enables evaluation from surface-level disambiguation to semantic retrieval, cross-lingual alignment, and interpretable idiom meaning.

## VI. Experiments and Results

This section presents experimental results across the four IdiomX benchmark tasks, evaluating classification, retrieval, cross-lingual alignment, and idiom interpretation. We report standard classification and retrieval metrics to assess how different modeling approaches handle idiomatic language understanding across progressively more challenging settings.

We begin with Task 1 as the foundational classification benchmark, followed by retrieval and interpretation tasks that evaluate increasingly deeper semantic reasoning.

### A. *Task 1: Idiom Detection*

#### 1. *Problem Definition*

Idiom detection evaluates whether an expression in context is used figuratively or literally, a classical challenge in figurative language understanding [1], [2]. We formulate this as a binary classification task where each input sentence is assigned one of two labels (Idiomatic, Literal).

This task measures contextual disambiguation ability and is important for downstream applications such as semantic parsing, machine translation, and dialogue systems.

#### 2. *Experimental Setup*

We evaluate three increasingly powerful contextual models:

- **TF-IDF + Logistic Regression** as a lexical baseline
- **DistilBERT [29]** as lightweight contextual transformer baseline.
- **RoBERTa [15]** as a stronger contextual encoder.

The dataset is nearly balanced (≈50% idiomatic / 50% literal) and split into train, validation, and test sets using stratified sampling.

All transformer models are fine-tuned using a maximum sequence length of 128 tokens, batch size 8, learning rate 2e-5, and 3 training epochs. Model selection is based on validation F1-score.

Performance is evaluated using Accuracy, Precision, Recall, and F1-score.

#### 3. *Results*

Table 2 presents idiom detection results.

| Model | Accuracy | F1-score |
|---|---|---|
| TF-IDF + Logistic Regression | 0.865 | 0.865 |
| DistilBERT [29] | 0.919 | 0.919 |
| RoBERTa [15] | 0.926 | 0.926 |

Table 2: Model Performance on Idiom Detection Task

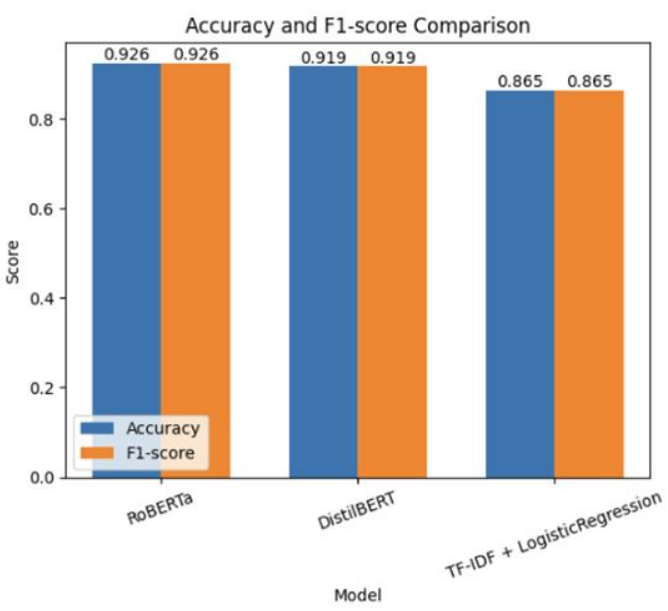

Figure 6: Idiom detection performance across models

Transformer-based models substantially outperform the lexical baseline, confirming the importance of contextual representation for modeling idiomatic meaning. RoBERTa [15] achieves the best performance, reaching 92.6% accuracy and 0.926 F1-score, demonstrating strong contextual disambiguation capability.

DistilBERT [29] provides a strong efficiency–performance trade-off, while RoBERTa [15] yields the strongest overall results.

#### 4. *Error Analysis and Calibration*

Despite strong overall performance, several recurring failure modes remain:

(1) Limited Context

Short or context-poor sentences can obscure whether an expression is literal or figurative.

(2) Idiom Bias

Frequent idioms may be incorrectly predicted as figurative even in literal contexts.

False positives are slightly more frequent than false negatives, reflecting mild idiom bias.

(3) Semantic Ambiguity

Some expressions permit both literal and idiomatic interpretations and require fine-grained contextual reasoning.

(4) Implicit Meaning

Models remain challenged when figurative meaning is only indirectly implied.

Confidence analysis and confusion patterns further show that errors often occur in highly ambiguous cases, with occasional overconfident incorrect predictions.

Representative calibration and error-analysis results are shown in Figures 7–11.

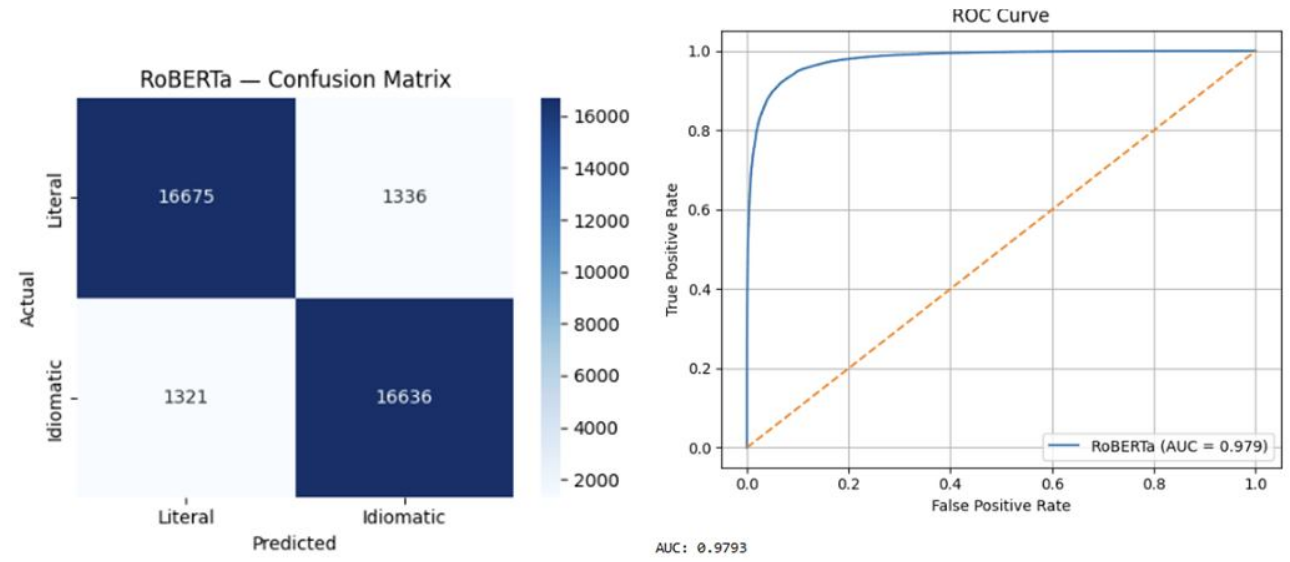

Figure 7: Confusion matrix for RoBERTa on the test set.

Figure 8: ROC curve for RoBERTa (AUC = 0.979).

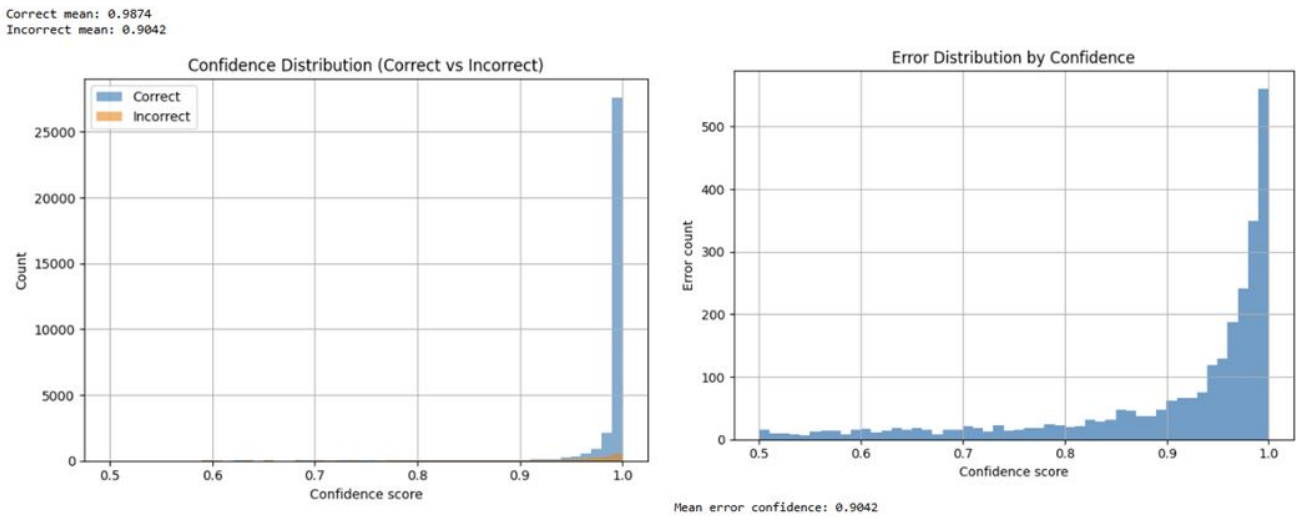

Figure 9: Confidence distribution for correct vs incorrect predictions.

Figure 10: Confidence distribution of prediction errors.

| | example | y_true | y_pred | confidence |
|---|---|---|---|---|
| 14488 | At first blush, the committee's reputation gai... | 0 | 1 | 0.999653 |
| 22463 | Oh sure, that book really blew her back out wi... | 0 | 1 | 0.999636 |
| 148856 | she opened the world like an oyster | 0 | 1 | 0.999608 |
| 132339 | My boss just slopped out a whole month's work ... | 0 | 1 | 0.999604 |
| 124023 | The CEO ran off with the company's confidentia... | 0 | 1 | 0.999599 |
| 170083 | Miner: If I strike gold here, it'll change my ... | 0 | 1 | 0.999594 |
| 81103 | Just saw my neighbors having a wild party—wish... | 0 | 1 | 0.999588 |
| 48052 | OMG! That concert was wow-worthy, add an excla... | 0 | 1 | 0.999564 |
| 118990 | Oh sure, she's a real race queen—she won the n... | 0 | 1 | 0.999542 |
| 136430 | The university's event stood us up by unexpect... | 0 | 1 | 0.999542 |

Figure 11: Examples of high-confidence errors.

#### 5. *Qualitative Examples*

Representative examples illustrate both strengths and limitations:

**Correct:**
"He finally kicked the bucket after a long illness."
Prediction: Idiomatic

**Failure:**
"He kicked the bucket lightly to move it aside."
Prediction: Idiomatic (gold: Literal).

These examples reflect the broader error patterns observed in quantitative analysis.

*6. Summary*

Results show that idiom detection benefits substantially from contextual transformer models, with RoBERTa [15] achieving the strongest performance. While detection accuracy is high, remaining errors highlight persistent challenges involving ambiguity, implicit meaning, and idiom memorization bias.

Overall, Task 1 establishes a strong classification benchmark within IdiomX and provides a foundation for the retrieval and interpretation tasks that follow.

### *B. Task 2: Context-to-Idiom Retrieval*

Task 2 evaluates whether a model can recover the correct canonical idiom from an English contextual sentence used idiomatically. We formulate this as a closed-set retrieval problem over a fixed idiom inventory:

$$y^{*} = \arg\max_{y \in I} score(x, y)$$

where $x$ is the input context, $I$ is the idiom inventory, and score$(x, y)$ is the retrieval score assigned to idiom $y$.

Unlike idiom detection, this task evaluates semantic retrieval rather than classification, requiring both contextual understanding and precise idiom ranking.

#### 1. Benchmark Construction

The benchmark is derived from the IdiomX high-quality subset using only examples that are:

(1) idiomatic in usage
(2) validation-approved
(3) high semantic quality

The resulting benchmark contains 54,469 examples and 13,803 canonical idioms.

A query-level split is used to avoid sentence leakage while allowing the same idiom to appear across splits in different contextual realizations.

| Setting | Value |
|---|---|
| Total examples | 54,469 |
| Idiom bank size | 13,803 |
| Train examples | 43,601 |
| Test examples | 10,868 |
| Split strategy | Query-level |
| Sentence leakage across splits | None |

Table 3: Task 2 benchmark statistics and split configuration.

#### 2. Retrieval Architecture

We evaluate four progressively stronger retrieval settings.

- **MiniLM Dense Retrieval** using sentence embeddings [28]
- **Hybrid Retrieval** combining MiniLM and BM25 [26]
- **Hybrid + Cross-Encoder Reranker** [28]
- **Hybrid + Fine-Tuned Reranker,** using task-specific hard-negative training.

This staged architecture follows a standard retrieve-then-rerank paradigm where early stages maximize recall and later stages refine ranking precision [17], [27], [29].

Performance is evaluated using:

1. Top-1 Accuracy
2. Top-3 Accuracy
3. Top-5 Accuracy
4. Mean Reciprocal Rank (MRR)

#### 3. Results

Table 4 reports retrieval performance.

| Model | Top-1 | Top-3 | Top-5 | MRR |
|---|---|---|---|---|
| MiniLM | 0.640 | 0.748 | 0.787 | 0.707 |
| Hybrid (MiniLM + BM25) | 0.761 | 0.857 | 0.882 | 0.817 |
| Hybrid + reranker | 0.838 | 0.901 | 0.906 | 0.869 |
| Hybrid + Fine-Tuned Reranker | 0.885 | 0.909 | 0.909 | 0.897 |

Table 4: Main Task 2 retrieval results on the closed-set benchmark.

Results show monotonic improvement as retrieval sophistication increases.

MiniLM [27] provides a strong dense retrieval baseline (Top-1 = 0.640), while hybrid retrieval substantially improves performance by combining semantic and lexical signals.

Adding reranking yields further gains, with the fine-tuned reranker achieving the best performance:

- Top-1**: 0.885**
- Top-3**: 0.909**
- Top-5**: 0.909**
- MRR**: 0.897**

These results confirm that idiom retrieval benefits from combining dense retrieval, lexical matching, and task-specific reranking [17], [27], [29].

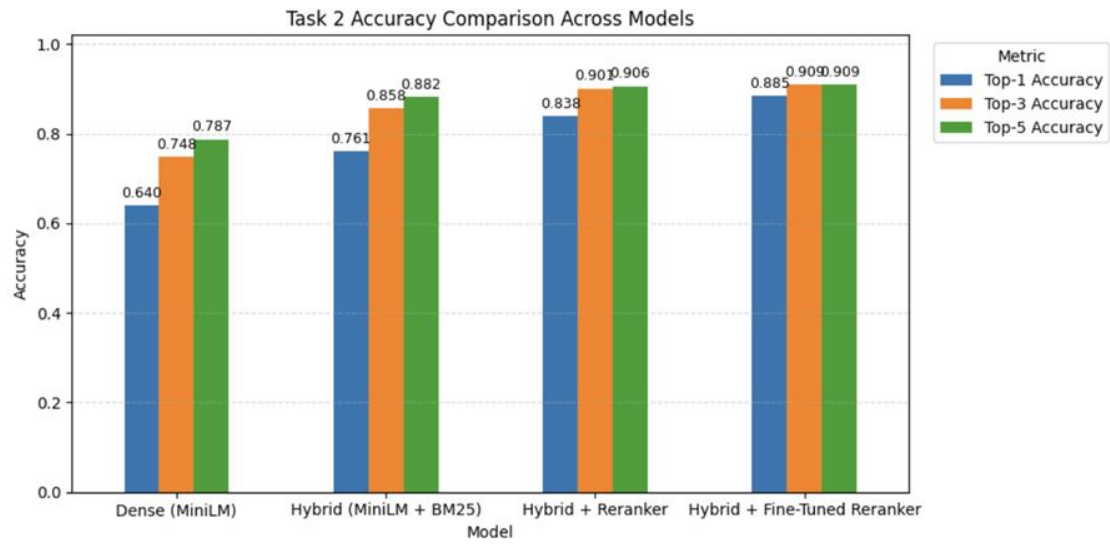

Figure 12: Performance comparison of Task 2 retrieval models across Top-1, Top-3, Top-5, and MRR.

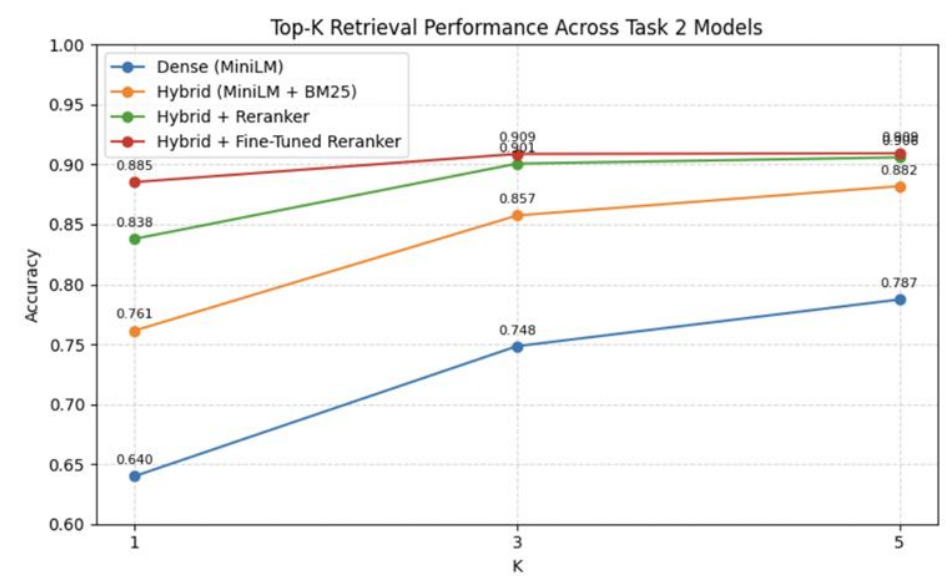

Figure 13: Top-$K$ retrieval curves for Task 2 under the closed-set retrieval setting.

Table 5 reports absolute gains over the dense retrieval baseline.

| Metric | Hybrid Gain | Hybrid + Reranker | Fine-Tuned Reranker |
|---|---|---|---|
| Top-1 Accuracy | +0.121 | +0.198 | +0.245 |
| Top-3 Accuracy | +0.109 | +0.153 | +0.161 |
| Top-5 Accuracy | +0.095 | +0.119 | +0.122 |
| MRR | +0.110 | +0.162 | +0.190 |

Table 5: Absolute improvement over the MiniLM dense retrieval baseline.

Three main findings emerge:

1. Hybrid retrieval significantly improves recall.
2. Reranking improves precision among top candidates.
3. Fine-tuning provides the largest gains for semantically similar idioms.

### 4. Error Analysis

Despite strong performance, approximately 11.5% of test queries remain incorrectly ranked.

Representative errors fall into four categories:

1. **Semantically Similar Idioms**
   Near-miss predictions where plausible alternatives outrank the gold idiom.
2. **Frequency Bias**
   Common idioms may be preferred over less frequent but correct expressions.
3. **Surface Lexical Bias**
   Strong lexical overlap can sometimes override deeper semantic matching.
4. **Ambiguous and Long-Tail Cases**
   Short contexts and rare idioms remain challenging.

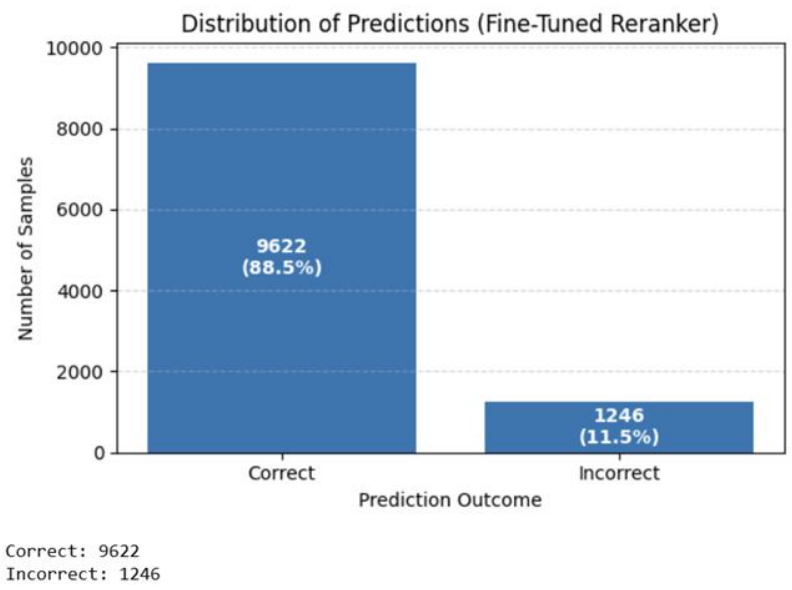

Figure 14: Distribution of correct and incorrect predictions for Task 2 on the test set.

These findings suggest that remaining errors stem primarily from ambiguity, frequency effects, and long-tail idioms.

### 5. Qualitative Examples

Representative examples from Table 6 illustrate both successful retrieval and common failure modes.

| Query (shortened) | True Idiom | Top-1 | Top-3 (shortened) | ✓ |
|---|---|---|---|---|
| "Bone apple tea… bon appétit" | bon appétit | throw a bone to | throw a bone…; wise apple; contrary… | ✗ |
| "…finally broke the ice…" | break the ice | break the ice | break the ice; ease tension; start talking | ✓ |

Table 6: Representative qualitative examples from Task 2

Correct predictions typically occur when contextual cues strongly signal the idiomatic meaning, while failures often involve semantically plausible but incorrect alternatives.

### 6. Summary

Task 2 shows that idiom understanding can be effectively modeled as a retrieval problem, but high performance requires more than semantic embeddings alone.

Hybrid retrieval, reranking, and task-specific fine-tuning substantially improve idiom recovery, with the fine-tuned reranker achieving the strongest performance.

Overall, Task 2 establishes a strong retrieval benchmark for idiomatic semantic understanding.

## *C. Task 3: Arabic Context-to-Idiom Retrieval*

Task 3 evaluates cross-lingual idiom retrieval, where the goal is to recover the correct canonical English idiom from an Arabic contextual sentence. Similar to Task 2, we formulate this as a closed-set retrieval problem over a fixed idiom inventory, but in a cross-lingual setting where the query is Arabic and the target idiom is English.

Given an Arabic query $x$, prediction is defined as:

$$y^* = \arg\max_{i \in I} sim(x, i)$$

where $\text{sim}(x, i)$ is cosine similarity [5] between the Arabic context embedding and candidate idiom embedding in a shared multilingual representation space.

### *1. Formulation*

*Similarity is computed using cosine similarity [5]:*

$$sim(x, i) = \frac{h_x \cdot h_i}{\|h_x\| . \|h_i\|}$$

### *2. Experimental Setup*

The final benchmark uses idiomatic examples only, where the Arabic query is taken from idiom-in-context Arabic examples and the retrieval target is the canonical English idiom. To preserve a realistic fixed candidate inventory while testing contextual generalization, we adopt example-level splitting rather than idiom-level splitting.

Candidate idioms are represented using multiple Arabic semantic views, including contextual examples and aligned meanings, encoded in a shared multilingual embedding space.

We evaluate three dense retrieval models of increasing capability:

- Multilingual MiniLM baseline [27]
- Multilingual E5 (zero-shot) [30]
- Fine-tuned multilingual E5 [30]using contrastive learning

Performance is measured using Top-1 and Top-5 retrieval accuracy.

### *3. Quantitative Results*

Performance is evaluated using standard ranking metrics, including Top-1 and Top-5 accuracy. Top-$K$ accuracy measures whether the correct idiom appears within the top $K$retrieved candidates.

Table 7 presents retrieval results.

| Model | Embedding Dim | Top-1 | Top-5 |
|---|---|---|---|
| MiniLM-L12-v2 [27] | 384 | 0.159 | 0.254 |
| E5-base (zero-shot) | 768 | 0.191 | 0.310 |
| E5-base (finetuned) | 768 | 0.578 | 0.761 |

Table 7: Retrieval performance on Task 3.

Results show consistent improvement across models. Generic multilingual embeddings yield limited retrieval performance, while zero-shot E5 [30] provides only modest gains. The strongest results are achieved through task-specific contrastive fine-tuning, which substantially improves semantic alignment between Arabic contexts and English idioms.

These results confirm that cross-lingual idiom retrieval is not merely a semantic similarity problem but requires targeted supervision and fine-grained multilingual representation learning.

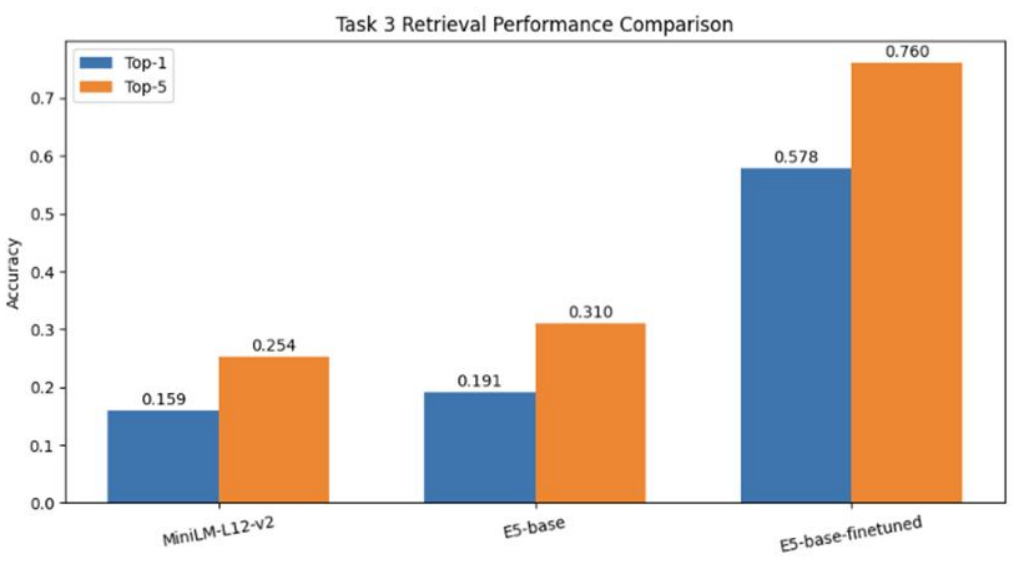

Figure 15: Model Retrieval Performance Comparison.

### 4. *Error and Calibration Analysis*

Error analysis reveals three major outcomes:

- Top-1 correct: 57.8%
- Correct idiom retrieved but misranked: 18.3%
- Fully incorrect: 23.9%

As shown in *Figure 16*, many failures arise from ranking errors rather than retrieval failures, since the correct idiom often appears within Top-5 candidates.

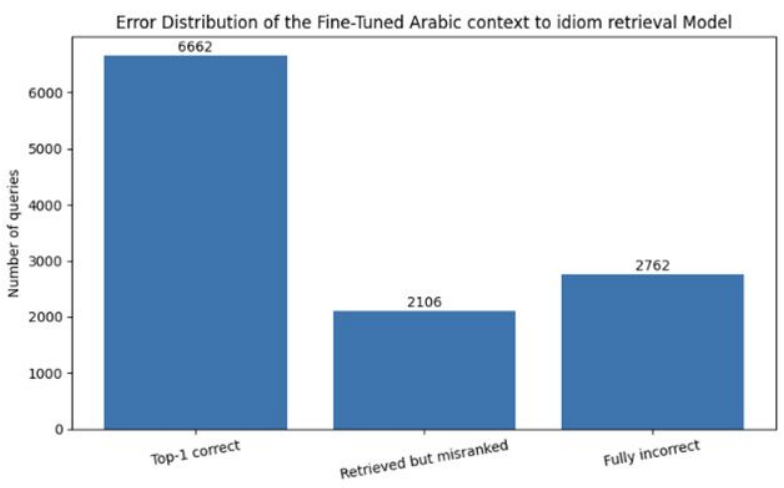

Figure 16: Error distribution across prediction outcomes.

Similarity and threshold analysis (*Figure 17*) further show that prediction accuracy increases consistently with similarity confidence, suggesting cosine similarity provides a meaningful calibration signal.

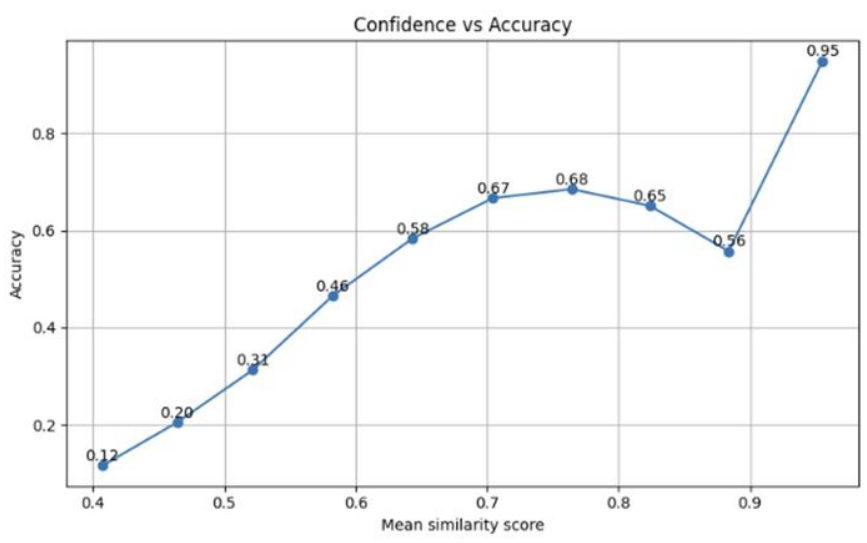

Figure 17: Accuracy as a function of similarity score.

Recall@K analysis (*Figure 18*) shows rapid gains at small values of $K$, indicating that the model often retrieves the correct semantic neighborhood even when ranking remains imperfect.

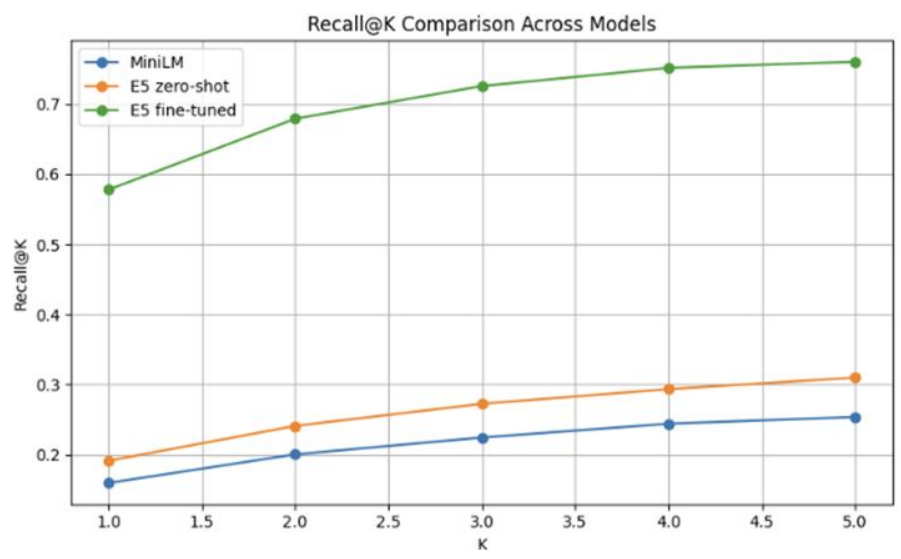

Figure 18: Recall@K performance across models.

Representative failure cases reveal that errors frequently involve semantically close idioms within dense meaning clusters, suggesting that future gains may come primarily from improved ranking precision and hard-negative training.

### 5. *Qualitative Cross-Lingual Analysis*

To better understand model behavior, we present representative examples illustrating both successful predictions and typical failure cases.

The model performs well when the Arabic context provides clear semantic signals. For example:

- Arabic: "انتهى كل شيء في النهاية"
  Prediction: *come to an end*
- Arabic: "شارك همومك مع الآخرين لتخفيفها"
  Prediction: *a problem shared is a problem halved*

In these cases, the model successfully captures the underlying meaning and maps it to the appropriate idiomatic expression.

However, failure cases reveal consistent challenges:

- Arabic: "يجب أن تبطئ كلامك حتى يفهمك الجميع"
  Prediction: *please speak more slowly*
  Target: *slow down*

Here, the model retrieves a semantically related phrase rather than the canonical idiom, indicating sensitivity to surface-level similarity.

### 6. *Summary*

Task 3 establishes a challenging benchmark for cross-lingual figurative retrieval. Results show that multilingual pretrained embeddings alone are insufficient, while task-specific fine-tuning substantially improves performance.

Although retrieval coverage is strong, remaining errors largely stem from ranking ambiguity among semantically similar idioms rather than complete retrieval failure. These findings highlight cross-lingual idiom retrieval as both a retrieval and semantic alignment problem, and suggest promising directions through ranking optimization and richer supervision.

## D. *Task 4: Idiom Interpretation*

While Tasks 2 and 3 focus on retrieving canonical idioms from contextual input, Task 4 addresses a complementary and less explored problem: interpreting idioms by retrieving their intended meaning, closely related to figurative semantic interpretation and multilingual meaning retrieval [3], [5], [17].

Given an idiom or idiomatic sentence, the goal is to recover its semantic interpretation and aligned multilingual meaning representation. Unlike idiom detection or idiom retrieval, this task evaluates whether a model can move from recognizing figurative expressions to explaining them.

We formulate Task 4 as a semantic retrieval problem over a meaning bank, where each idiom is associated with aligned multilingual meaning representations in English, Arabic, and French.

### 1. Task Formulation

Given an idiom query $x$, prediction is defined as:

$$m^* = \arg\max_{m \in M} sim(x, m)$$

where $M$ denotes the meaning inventory and $sim(\cdot)$ is cosine similarity in a shared embedding space.

Each retrieved result returns:

- Canonical idiom
- English meaning
- Arabic meaning
- French meaning (when available)

This formulation evaluates semantic grounding rather than idiom recovery alone.

### 2. Experimental Setup

We construct the meaning retrieval bank from aligned semantic fields in IdiomX, combining:

- canonical idioms
- idiom-level meanings
- contextual meaning paraphrases
- multilingual aligned meanings

We evaluate dense semantic retrieval using multilingual sentence embeddings and retrieval-based semantic matching methods inspired by Sentence-BERT retrieval and dense ranking approaches [17], [27], [28].

Performance is evaluated using Top-1 and Top-5 retrieval accuracy together with Mean Reciprocal Rank (MRR), following the retrieval setup used in Tasks 2 and 3.

### 3. Quantitative Results

Performance is evaluated using Top-1, Top-3, Top-5 retrieval accuracy and Mean Reciprocal Rank (MRR), following the retrieval protocol used in Tasks 2 and 3.

Table 8 reports Task 4 results.

| Model | Top-1 | Top-3 | Top-5 | MRR |
|---|---|---|---|---|
| Dense Retriever (MiniLM) | 0.355 | 0.462 | 0.501 | 0.411 |
| Hybrid Retrieval | 0.659 | 0.710 | 0.714 | 0.684 |
| **Hybrid + Reranker** | **0.674** | **0.713** | **0.716** | **0.693** |

Table 8: Idiom interpretation performance on Task 4.

Results show consistent improvement as retrieval complexity increases. Dense semantic retrieval provides a strong baseline, while combining semantic retrieval with lexical matching substantially improves idiom meaning recovery. Adding reranking yields further gains, with the hybrid reranked model achieving the strongest overall performance.

Hybrid + reranking improves Top-1 accuracy from 35.5% to 67.4%, nearly doubling dense retrieval performance

Figure 19 illustrates model progression across retrieval settings.

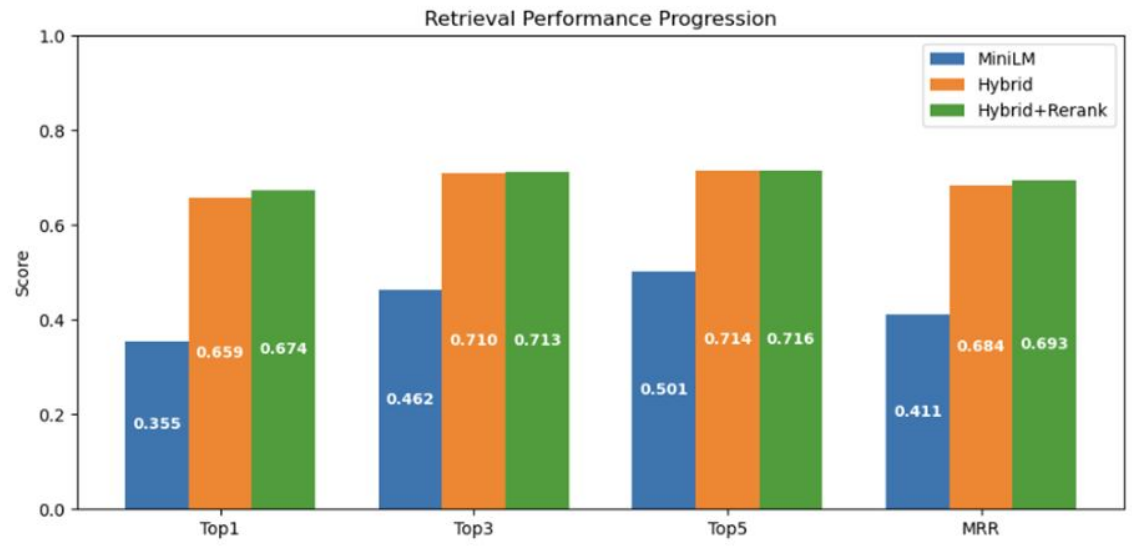

- *Figure 19:* Model comparison across interpretation systems

.

Figure 20 shows Top-K retrieval curves. Rapid gains at small values of K indicate that even when the correct meaning is not ranked first, it frequently appears among the top retrieved semantic candidates.

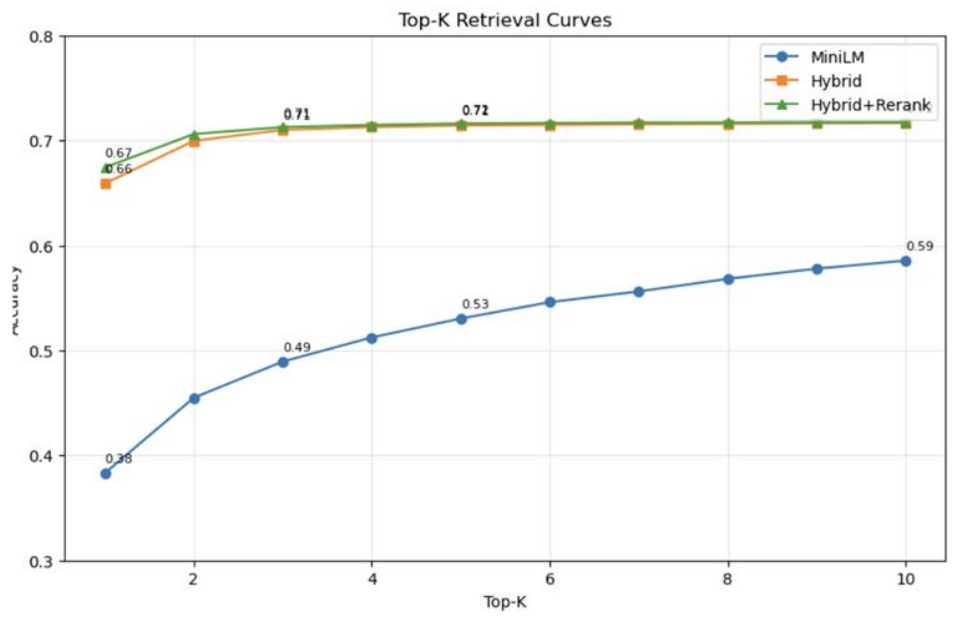

*Figure 20: Top-K retrieval performance curves for Task 4.*

These results confirm that idiom interpretation benefits from the retrieval-then-rerank paradigm widely used in semantic retrieval literature [17], [27], [29], while introducing a benchmark focused on semantic grounding rather than idiom recovery alone.

### 4. Error Analysis

Remaining failures largely fall into three categories:

1. Short or ambiguous idioms with underspecified context
2. Lexical confusion between semantically related paraphrases
3. Semantic confusion among near-neighbor idiomatic meanings

Figure 21 shows the distribution of these failure modes.

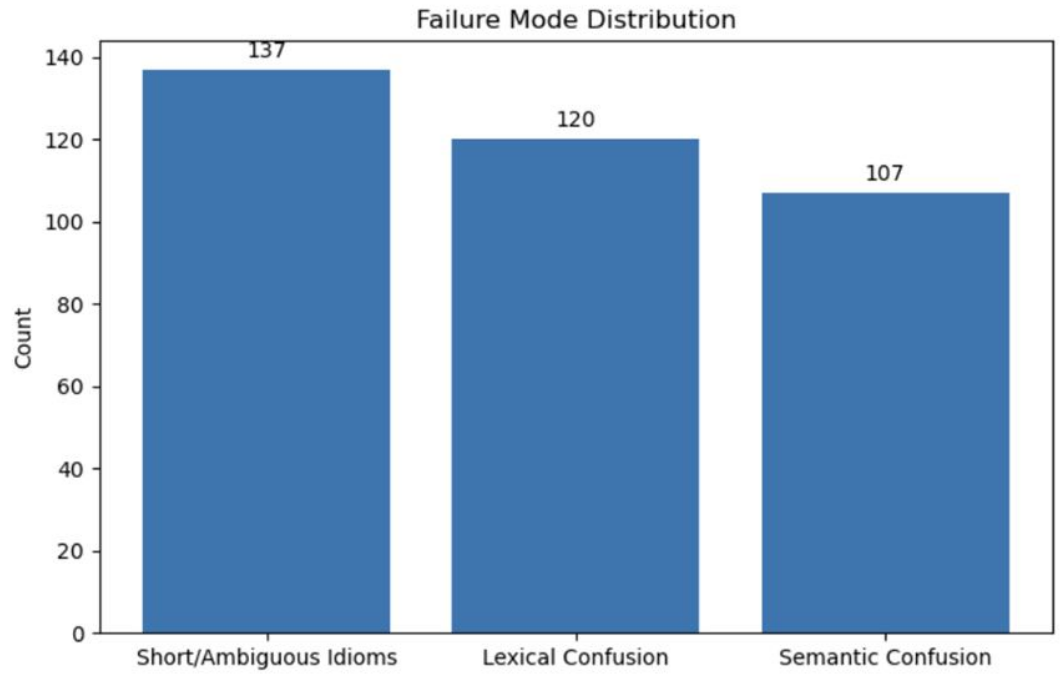

*Figure 21:* distribution of these failure modes.

Unlike retrieval failures in Tasks 2–3, many Task 4 errors involve semantically plausible interpretations rather than entirely incorrect predictions, suggesting ranking precision rather than representation quality is often the main limitation.

### *5. Qualitative Examples*

Representative examples illustrate the interpretation capability:

**Input:** *spill the tea*
Retrieved meaning: reveal gossip or personal secrets
Arabic: كشف الشائعات أو الأسرار
French: révéler des potins ou des secrets

**Input:** *break the ice*
Retrieved meaning: reduce social tension and start conversation
Arabic: كسر الجمود وبدء الحديث
French: briser la glace socialement

Failure cases generally involve semantically related interpretations rather than complete meaning mismatches, for example retrieving *reveal confidential information* instead of *reveal gossip*, showing high semantic overlap but imperfect ranking.

### *6. Discussion*

Task 4 extends idiom understanding beyond detection and retrieval into semantic interpretation, introducing an explainability-oriented benchmark dimension absent from prior idiom benchmarks [1]–[5].

This task is particularly relevant for:

- language learning systems
- multilingual semantic search
- explainable NLP systems
- machine translation support
- human-AI interaction

By grounding idioms in multilingual meanings, Task 4 complements the retrieval tasks and broadens IdiomX from a retrieval benchmark into a more general figurative language understanding framework.

### *7. Summary*

Results show that idiom interpretation can be effectively modeled as semantic retrieval, with hybrid retrieval and reranking achieving strong performance.

Task 4 introduces a new benchmark dimension focused on semantic grounding and interpretability, extending IdiomX beyond idiom recognition toward explainable figurative language understanding.

| Task | Best Model | Main Result |
|---|---|---|
| Detection | RoBERTa [15] | 92.60% |
| Retrieval | Hybrid+FT reranker | 88.5% Top-1 |
| Arabic Retrieval | Fine-Tune E5 [30] | 57.8% Top-1 |
| Interpretation | Hybrid+Reranker | 67.4% Top-1 |

Table 9: benchmark summary

## VII. Discussion

The experimental results demonstrate that idiom understanding is not a single problem, but a spectrum spanning contextual disambiguation, semantic retrieval, cross-lingual alignment, and idiom interpretation.

Three key findings emerge.

First, idiom detection primarily depends on contextual modeling rather than surface lexical cues, confirming observations from prior figurative language benchmarks that idiomatic meaning cannot be reliably inferred compositionally [1], [2], [4].

Second, retrieval tasks (Tasks 2–4) show that idiom understanding is fundamentally a semantic ranking problem. While dense retrieval models capture relevant semantic neighborhoods, strong performance requires hybrid retrieval and reranking mechanisms for fine-grained idiom selection and meaning grounding, consistent with prior semantic retrieval findings [17], [27], [29].

Third, cross-lingual retrieval and idiom interpretation remain challenging due to ambiguity among semantically related idioms and multilingual meaning variation. These findings align with broader challenges reported in multilingual representation learning [5], [6].

A broader pattern across tasks is that hybrid architectures consistently outperform single-paradigm approaches, indicating that figurative language understanding benefits from combining contextual reasoning, structured retrieval, and semantic grounding within a unified framework.

Task 4 further extends this perspective by introducing interpretability as an evaluation dimension, moving beyond idiom recognition toward explainable figurative language understanding.

## VIII. Applications

IdiomX supports a broad range of applications involving figurative and multilingual language understanding, including:

- idiom detection and disambiguation
- context-to-idiom retrieval
- cross-lingual idiom mapping
- idiom interpretation and semantic explanation
- multilingual semantic search
- language learning and intelligent tutoring
- machine translation support
- explainable NLP and human–AI interaction

Beyond benchmarking, IdiomX can support evaluation and development of transformer models, retrieval systems, and large language models on non-compositional language phenomena that remain challenging for current NLP systems [6], [13].

## IX. Limitations

Despite its scale and breadth, IdiomX has several limitations.

First, portions of the dataset rely on controlled LLM-based enrichment, which may introduce stylistic regularities or residual synthetic biases despite extensive validation.

Second, while multilingual coverage includes English, Arabic, and French, cross-lingual coverage remains limited relative to the broader diversity of idiomatic variation across languages and dialects.

Third, retrieval evaluation primarily uses ranking-based metrics (Top-K and MRR), which may not fully capture semantic equivalence among paraphrastic idiom interpretations, a known challenge in retrieval evaluation [17].

Finally, although Task 4 introduces interpretability-oriented evaluation, assessing explanation quality remains an open problem requiring richer human-centered evaluation.

Future work includes broader human validation, dialectal expansion, richer semantic evaluation protocols, and extension to additional languages.

## X. Conclusion

We introduced IdiomX, a large-scale multilingual benchmark for idiomatic language understanding, combining over 190K contextualized examples, 12K+ idioms, aligned multilingual semantics, and a reproducible multi-stage construction pipeline.

We further presented a unified four-task benchmark covering idiom detection, context-to-idiom retrieval, Arabic-to-English idiom retrieval, and idiom interpretation, enabling evaluation across classification, retrieval, cross-lingual alignment, and semantic grounding.

Experiments show that contextual transformers substantially improve idiom detection [13], [14], while hybrid retrieval and reranking approaches significantly strengthen monolingual and multilingual idiom retrieval [17], [27], [29]. Results from Task 4 further highlight the importance of semantic grounding and interpretability as complementary dimensions of figurative language evaluation.

Overall, IdiomX positions idiom understanding as a multi-dimensional problem requiring contextual reasoning, structured retrieval, multilingual representation, and explainable semantic interpretation.

We hope IdiomX provides a scalable foundation for future research in figurative language, multilingual NLP, and idiom-aware language models.

## XI. Reproducibility and Public Resources

To support transparency, reproducibility, and community reuse, all resources associated with IdiomX are publicly released, including the dataset, construction pipeline, benchmark implementations, trained models, and the unified interactive platform IdiomX Studio [31].

### A. Dataset Release

The full IdiomX dataset is publicly available through Hugging Face:

**Dataset:**
https://huggingface.co/datasets/aymansharara/IdiomX

The release includes all dataset configurations, benchmark splits, multilingual semantic fields, and enriched annotations used in this work, supporting all four benchmark tasks.

### B. Code and Experimental Pipelines

The complete dataset construction pipeline and experimental code are publicly available through GitHub:

**Dataset construction pipeline:**
https://github.com/aymanshar/idiomx-dataset

**Models and benchmark implementations:**
https://github.com/aymanshar/IdiomX

These repositories include:

• multi-stage dataset construction scripts
• LLM enrichment and validation pipelines
• training and evaluation code for all benchmark tasks
• released model artifacts and benchmark resources
• instructions for reproducing experiments and figures reported in this paper

Together, these resources support full end-to-end reproducibility of the reported results.

### C. IdiomX Studio: Unified Benchmark Interface

To support interactive exploration of the benchmark, we provide IdiomX Studio, a unified demonstration platform integrating all four benchmark tasks within a single interface:

**IdiomX Studio (Hugging Face Space)**
https://huggingface.co/spaces/aymansharara/idiomx-studio

The platform supports:

• Task 1: Idiom Detection
• Task 2: Context-to-Idiom Retrieval
• Task 3: Arabic Context-to-Idiom Retrieval
• Task 4: Idiom Interpretation

Unlike task-specific demos, IdiomX Studio provides a unified interface for exploring model behavior across classification, retrieval, cross-lingual alignment, and semantic interpretation within a single system.

The platform also integrates released benchmark models, retrieval artifacts, and interactive examples, enabling qualitative inspection of predictions and practical evaluation beyond static benchmark metrics

### D. Reproducibility Statement

All experiments use publicly released data splits, open-source implementations, fixed benchmark protocols, and released model artifacts, enabling independent replication and future extension of this work.